\documentclass[%
preprint,
 amsmath,amssymb,
 aps,
]{revtex4-2}

\usepackage{listings}

\usepackage{enumitem}
\usepackage{color}
\usepackage{mathtools}
\usepackage{graphicx}
\usepackage{dcolumn}
\usepackage{bm}

\begin{document}


\title{Function Decomposition Tree with Causality-First Perspective and Systematic Description of Problems in Materials Informatics}
\author{Hiori Kino}
\affiliation{Research and Services Division of Materials Data and Integrated System,
National Institute for Materials Science, 1-1 Namiki, Tsukuba, Ibaraki 315-0044, Japan}
\author{Hieu-Chi Dam}
\affiliation{Japan Advanced Institute of Science and Technology,
1-1 Asahidai, Nomi, Ishikawa 923- 1292, Japan}
\affiliation{
International Center for Synchrotron Radiation Innovation Smart Cross Fertilization Division, 
Tohoku University, Katahira 2-1-1, Aoba-ku, Sendai, Miyagi, 980-8577, Japan}
\author{Takashi Miyake}
\affiliation{Research Center for Computational Design of Advanced Functional Materials,
National Institute of Advanced Industrial Science and Technology,
1-1-1 Umezono, Tsukuba, Ibaraki 315-8568, Japan}
\author{Riichiro Mizoguchi}
\affiliation{Japan Advanced Institute of Science and Technology,
1-1 Asahidai, Nomi, Ishikawa 923- 1292, Japan}

\date{\today}

\begin{abstract}
As interdisciplinary science is flourishing because of materials informatics and additional factors; a systematic way is required for expressing knowledge and facilitating communication between scientists in various fields. A function decomposition tree is such a representation, but domain scientists face difficulty in constructing it. Thus, this study cites the general problems encountered by beginners in generating function decomposition trees and proposes a new function decomposition representation method based on a causality-first perspective for resolution of these problems. The causality-first decomposition tree was obtained from a workflow expressed according to the processing sequence. Moreover, we developed a program that performed automatic conversion using the features of the causality-first decomposition trees. The proposed method was applied to materials informatics to demonstrate the systematic representation of expert knowledge and its usefullness.
\end{abstract}


\maketitle

\section{Introduction \label{sec:intro}}
In general, useful information can be obtained by relating and organizing the components (entities) of things with each other, and there are many examples in materials informatics, the area of application of this study. The most familiar example for material informatics is that of representing a molecule as a graph. By representing an entity (molecular structure) as a graph of molecular bonds \cite{REFSIMLES} and analyzing it using machine learning methods, it is possible not only to regress the physical properties but also to construct molecular generative models.\cite{REFDeepLearningGraph,REFMoleculerRepresentation,REFDesigningandUnderstandingLightDevice} In inorganic materials, the definition of neighboring atoms is ambiguous; however, by representing the 3D atomic structure as a graph, explanatory variables can be obtained from data by regressing the total energy \cite{REfLearningAtoms}. Explanatory variables can also be obtained from data using supervised learning of the time evolution of the glass simulation from the initial atomic positions.\cite{REFGlassySystems} 

The prediction of physical properties using graphs as basic data includes the use of knowledge graphs.\cite{REFMultipleMaterialsNN} In polymers, there is a high dependence of electrical conduction on impurities, therefore it is difficult to identify and quantify the type of impurity and the changes in electronic state caused by the impurity.\cite{REFDNAdoping} In such cases, graphs, which directly relate the dependence of preparation conditions to electrical conduction have been created, can perform missing problems to infer unmeasured conditions, values.

There is another stream of research using knowledge graphs. The relationship between entities itself is sometimes useful information. Studies on a citation map \cite{REFScienceMapping} have considered other entities, such as material properties, as well as the citation relationship.\cite{REFLiFePO4} One study aimed to understand the correlation of entities (properties) from the data \cite{REFPropnet} which makes it easy to understand the relationship between properties. 

Some studies use ontologies to organize relationships. Research in this direction includes work on inferring chemical reaction kinetics and reaction mechanisms\cite{REFOntoKin} which is similar to IBM Watson \cite{REFWatson}. Ontologies give concepts, attributes, and a network of relationships (TBox) and generate instances to embody the chemical or reaction of interest. Next, experimental data are taken in to create a database, and Q\&A (reasoning) is exactly the same as database searching, but the ontology (conceptual structure) allows for high-level searching while reasoning.

To reasonably infer and predict the knowledge of physical properties and mechanisms, it is, of course, necessary to accurately describe the relationships. In recent years, studies have aimed to obtain relations (summarizing knowledge) by natural language processing\cite{REFRelationExtraction} but the most accurate property values are still obtained by humans or modified by humans.\cite{REFPualingFile} Therefore, human assistance is still important for knowledge compilation. However, no one person knows everything in the domain, and even if one did, it would be impossible for one person to obtain all the values from the literature. To obtain a reasonable experimental physical property value from the literature, it is necessary to know, for example, the dimension of the values (scalar or vector), the type of values (median or not), and the unit in which the value can be expressed. In addition, due to the isotropic nature of crystals, there are cases in which an originally vector quantity is treated as a scalar quantity. These can be classified in the ontology, but generally we cannot assume everything in the conceptual hierarchy structure of the initial ontology; thus, we notice flaws in the hierarchy during data acquisition. Therefore, having knowledge of not only the results of the ontology but also the reasons why it was constructed allows us to summarize and share knowledge among research groups efficiently in a limited amount of time and budget.

Function decomposition tree can be used for this purpose. It has been proposed as a method to represent the relationships among entities as a diagram assuming that it is mainly used by humans to visualize the relationships among entities themselves for the purpose of conceptual design, understanding, and idea support, and their usefulness has been recognized.\cite{REFOntologicalModel} (Diagrams are defined as general visualization techniques that include the flowchart and the function decomposition tree.) Although the function decomposition tree was initially proposed for the decomposition of functions of artefacts, it can also be applied to the decomposition of (human) actions due to the similarity between the functions of artefacts and actions.\cite{REFCHARM} Therefore, the scope of the function decomposition trees includes not only the principle of the artificial device to be specifically executed, but also the procedures and calculation methods to solve the problem (algorithms) and the action.
The usefulness of diagrammatic representation on knowledge is also discussed in physics education. The relationship between reinterpretation of physics problems and the organization of physics knowledge has been investigated by means of diagrammatic representations for experts and novices. Experts and novices can start their problem representations from different problem categories, but the completeness of the representation depends on the domain knowledge{REFExpertsandNovices} suggesting that the difference in the domain knowledge is the main factor distinguishing experts and novices in their problem-solving abilities.\cite{REFCognitiveLoad} Perhaps for this reason, it is not effective for novices to create diagrammatic representations of even simple problem,\cite{REFTrainigStudent} but on the other hand, diagrammatic methods that convey expert knowledge to novices can also contribute significantly to improving their problem-solving abilities. For example, it is reported that novices can select almost the same methods as those of experts by using the function decomposition tree prepared by experts in nanomaterial domain.\cite{REFDesingSupporting} One might think that purpose-based decomposition algorithms, which the function decomposition trees use, requests high a cognitive load for humans, because it requires a large amount of memory when exactly solved by a computer. However, the use of external memory in the form of diagrammatic representations compiled in the form that humans can efficiently understand may help to reduce the cognitive load for humans greatly.\cite{REFCognitiveLoad}

The function decomposition is summarized in this paragraph.\cite{REFOntologicalModel} It is a visualization of an algorithm or a sequence of actions emphasizing what it does rather than how it runs.
\begin{itemize}
\item[i.1)] Contrary to ordinary chronological descriptions of sequence of actions, it explicates the intermediate goals to be achieved (purpose-based decomposition algorithm).
\item[i.2)] It makes the hidden intermediate goals explicit as well as final goals that even the actors (inventors of the algorithm) themselves are not aware of.
\item[i.3)] A single (sub)goal can be achieved in multiple ways. A decomposition tree allows us to draw them in a single tree so that one can capture the whole picture of a family of related procedures with many possible variants, which is enabled by decomposing a function into what to achieve from how to achieve. This contributes to the in-depth understanding of algorithms/procedures.
\item[i.4)] This is not an ontology, but it is one of the most valuable results derived from the ontology engineering research. Furthermore, it provides a "point of view" to "understand" algorithms ontologically (to explicate their essentials).
\end{itemize}
Thus, the following benefits are expected in practice. 
\begin{itemize}
\item[ii.1)] Even people who are not familiar with algorithms can understand the essence of the algorithms.
\item[ii.2)] Multiple algorithms sharing the same final goal can be superimposed on a single function decomposition tree (general purpose function decomposition tree), so that mutual similarities and differences can be made explicit.
\item[ii.3)] It can predict the spillover effects of algorithm modifications, which is useful when improving procedures.
\end{itemize}

Function decomposition trees have been studied outside of the field where ontology-related research was originally conducted, for example, in collaboration with researchers who are experts in lifestyle research \cite{REFLifeStyle1,REFLifeStyle2,REFLifestyle3}. On the other hand, in fields where ontology-related research is minor, it is not easy for the domain experts to create function decomposition trees. Therefore, as stated in the first paragraph, this study is written assuming that the primary target audience is materials informatics researchers.

The function decomposition tree is one of the top-down methods, and another example is Concept Map.\cite{REFConceptMap} The latter differs from the former in that it uses verbs, prepositions, and specializations to decompose the relationships between concepts, and it has been proposed as an educational tool and used as a knowledge integration tool. The drawback of the top-down methods is that they require a lot of explanation and examples for proficiency, and it has been reported that the learning cost for creating the concept map is not low. Even with these drawbacks, the function decomposition tree and the concept map have been studied and have applied in other domains because they offer sufficient returns.

Function decomposition trees in this manuscript may seem obvious to experts who have already understood the algorithms in materials informatics. If the explanation by the function decomposition seems natural and plain to an expert, then a novice can expect to understand the important points in a logical manner, thus demonstrating the usefulness of function decomposition as an explanatory tool.

Some readers may think that the idea of the function decomposition tree is so-called means-ends analysis.\cite{REFHumanProblemSolving} However, the difference is that the latter is at the level of strategy or algorithm, whereas the former is based on the results of ontology engineering and gives concrete ways to realize it through a simple strategy of the function decomposition.

This study applies the function decomposition to the conversion algorithm from atomic descriptors to crystal descriptors in materials informatics, which suggests that there are conversion methods that have not been used so far. (It will be discussed in Section~\ref{sec:Functiondecompositionchains}) We believe that this research will help to demonstrate the usefulness of function decomposition of existing algorithms.

In this study, Section~\ref{sec:sequentialandpurposedecomp} briefly introduces the conventional function decomposition tree and highlights the problems that hinder its application for beginners. In Section~\ref{sec:verb}, we define the most basic verbs used in function decomposition in the field of materials science and materials informatics. In Section~\ref{sec:newfuncdeomp}, we present a novel function decomposition method that compensates for its shortcomings. In addition, the novel function decomposition method allows for direct mapping to a time-series description (sequential algorithm), and a simple program is introduced for this conversion. In Sections~\ref{sec:Applicationsystematicexplanation} and \ref{sec:Applicationsystematicexplanation}, function decomposition was applied for material search algorithms and a Python package. Section~\ref{sec:Discussionfunctiondecomposition} contains a discussion, and Section~\ref{sec:Discussionfunctiondecomposition}  summarizes this work.

\section{Sequential Algorithm and Purpose-Based Decomposition Algorithm\label{sec:sequentialandpurposedecomp}}

The basics of the conventional function decomposition trees described in \cite{REFOntologicalModel} are briefly explained in this section. First, devices, behaviors, and functions in sequential algorithms and actions are introduced.

\subsection{Devices, Behaviors, and Functions in Sequential Algorithms\label{sec:devicebehaviors}}

The sequential algorithm is described with its accompanying concepts using heat transfer as an example. The \textbf{device} (an artefact) is defined as entity that converts the input object(s) into output object(s). In the following, devices that exchange heat are considered concretely. Devices that use heat are useful; for example, materials with low thermal conductivity \cite{REFLowThermalConductivity} and high-performance thermoelectric materials that convert heat into electricity \cite{REFNewThermoelectricMaterial} are being explored through machine learning techniques. Radiant coolers{REFAdiativeCooling} which radiate heat directly with no object to exchange, are also attracting attention. The following paragraphs use an example of these heat exchanger to illustrate its behavior and function and related issues.

\begin{figure}
\includegraphics[width=12cm]{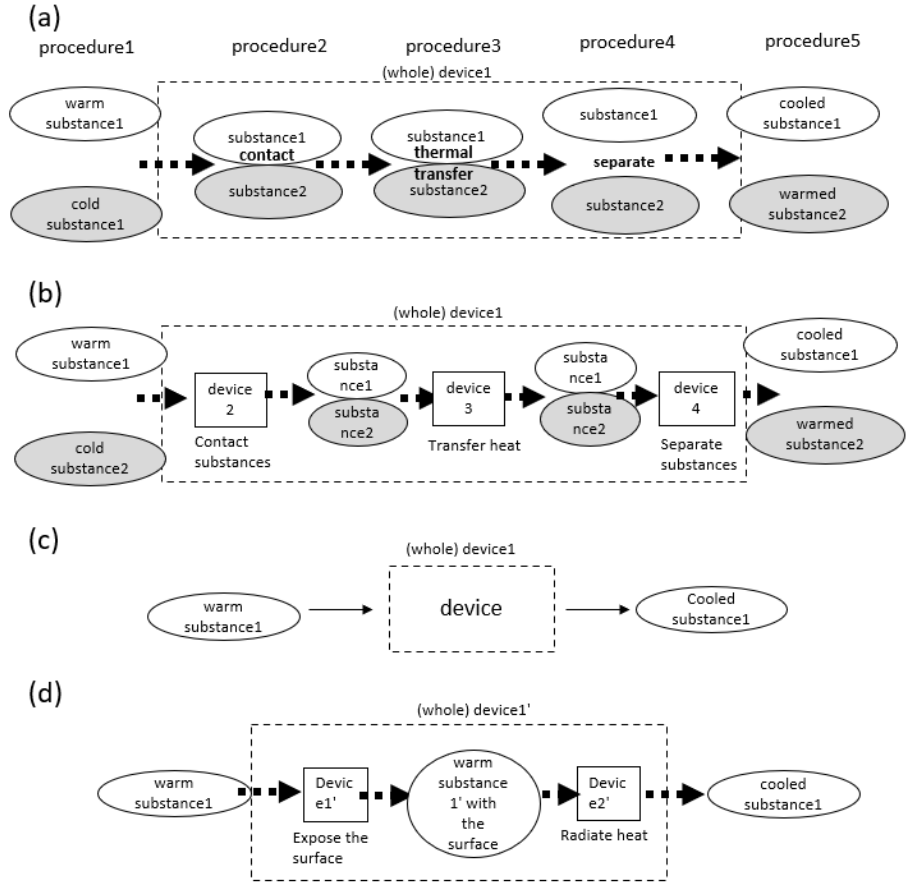}
\caption{
Figure~\ref{fig:heatExchanger}: (a) A contact heat exchanger with "warm substance 1 and cold substance 2" as input, "cold substance 1 and warm substance 2" as output, and its internal process procedures, which are converted to the object-attribute flowchart presented in (b). (c) depicts a device exhibiting a cooling function (converting warm material into cold material). (d) denotes a device and its internal processing procedures that cools via thermal radiation.
}
\label{fig:heatExchanger}
\end{figure}

A contact heat exchanger is depicted in Figure~\ref{fig:heatExchanger} (a), where the "warm substance 1 and cold substance 2" is the input and the "cold substance 1 and warm substance 2" is the output.
All pairs of inputs and outputs are termed as the \textbf{behavior} of the device action. Thus, the number of input and output pairs corresponds to that of the behaviors of the device. However, in practice, the behavior of the device will be represented by a pair of attentions. Thus, the behavior of the heat exchanger (device 1) shown in Figure~\ref{fig:heatExchanger} (a) is represented by the contact of the substance 1 and substance 2, which transfers heat and separates both the substances.

A device is the most basic entity that acts on the object(s) to alter the \textbf{object-attribute}(s) by converting the input to the output of the device.\cite{REFFBRL} Moreover, devices can be virtual. The entire process can be considered in terms of the sequential change in object-attribute(s), and the device can be interpreted as acting in the sequence: contact, transfer heat, and separate, as shown in Figure~\ref{fig:heatExchanger} (b). Consequently, the attributes of both substances become contacted substances, heat exchanged substances, and separated substances, respectively. (Although contact and heat transfer occur simultaneously, we aligned the sequence based on the cause-effect principle of the two processes, i.e., contact precedes heat transfer.) Hereafter, the diagram, such as Figure~\ref{fig:heatExchanger} (b), is referred to as the \textbf{object-attribute flowchart}.

The term function is defined as the role played by the behavior of the device under designer's intention.\cite{REFCharacteringFunctions} If the design intention of the device is to obtain a cold substance, the function would change warm substance 1 into cold substance 1, i.e., cool substance 1 as shown in Figure~\ref{fig:heatExchanger} (c).
The same cooling function of the various devices can be clearly discussed on the same footing by narrowing down the action of the device to its function. Another cooling device is described in the object-attribute flowchart as device 1' in Figure~\ref{fig:heatExchanger} (d), which exposes the surface of a material to radiate heat.

\section{Function decomposition Tree in Purpose-Based Decomposition Algorithm\label{sec:funtiondepintro}}

The way how to achieve the function considered in this section. Any artefact creation problem, such as device design, can be distinguished in three tiers: i) a (parent) function as a requirement, ii) a device that realizes and determines the behavior, and iii) furthermore, a (concrete internal) structure that realizes the behavior (inside the device) is determined.
Once a device is selected under the designer's intention, the behavior inside the device is a mechanism for realizing the function using a sequence of children functions. In addition, the device and its \textbf{implementation way} may be considered the same depending on the context, because a device embodies a single implementation, regardless of being conceptually different from other devices.
The conventional function decomposition was defined with the following two ways:
\begin{itemize}
\item[\#1.]	The parent function is decomposed into multiple child-functions via the implementation way. This may be done hierarchically as parent function $\longrightarrow$ parent implementation way $\longrightarrow$ child function $\longrightarrow$ child implementation way $\longrightarrow$ grandchild function $\longrightarrow$ grandchild implementation way $\longrightarrow$ and so forth. This is called the (function) \textbf{decomposition way}. The specific structure of the device has been described later.
\item[\#2.]	The substitution of a function with a specialized function positioned at a level lower than the (parent) function in the is-a hierarchy is termed as the \textbf{specialization way}. Although this operation is not "decomposition", it can be regarded as a type of decomposition, considering that the essential purpose of the function decomposition brings functions closer to the implementation level by reducing the granularity of the functions.
\end{itemize}

The two function decomposition methods are explained in Table~\ref{tag:method}. Note that the term "decomposition way" is used instead of the term "implementation way" to represent the context contrary to the object-attribute flowchart.

\begin{table}
\caption{Two function decomposition methods.}
 
\hrulefill
\begin{itemize}
\item[\#1.]  Decomposition by Decomposition Way 
\begin{itemize}
\item[1a.]  Select a decomposition way for the decomposition of a parent function.
\item[1b.]  Decompose into a series of child functions with a specific sequence of processes according to the decomposition scheme.
\end{itemize}
\item[\#2.]  Decomposition by Specialization Way
\begin{itemize}
\item[2a.]  Replace a parent function with a child function that is a specialization of the parent function (specialization via is-a relation).
\end{itemize}
\end{itemize} 

\hrulefill
\label{tag:method}
\end{table}
  
The function decomposition employs the black-box model to prohibit access to the behaviors inside the device. The access to the inside of the device is allowed only for child functions. Although the decomposition of the parent function into multiple children functions apparently allows us to access the inside of the device --- turning the \textbf{black box} to the \textbf{white box} --- this access is allowed at the function level rather than the behavior level. Thus, the black box model is maintained. The nested structure for black-boxing of schemes at higher level functions and the corresponding \textbf{white-boxing} with function decomposition of those schemes continues until the function decomposition has been exhausted.

\begin{figure}
\includegraphics[width=12cm]{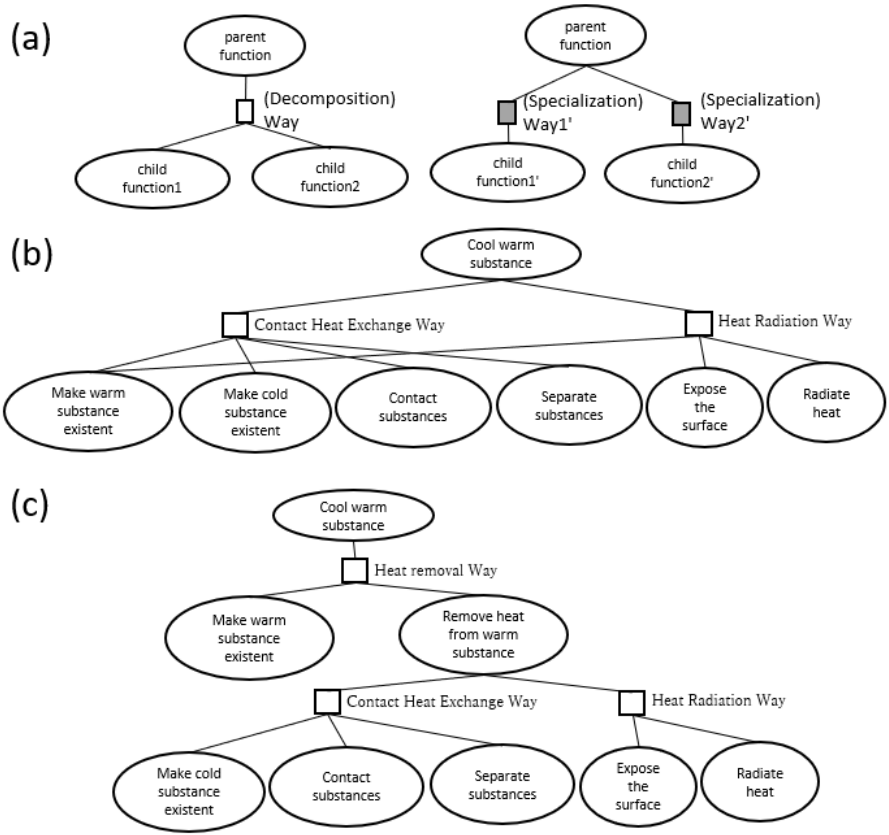}

\caption{
Figure~\ref{fig:firstFunctionDecompotion}: (a) The parent function is achieved in the decomposition way by processing child function 1 and child function 2 in sequence. In specialization, ways 1' and 2' are accomplished by child functions 1' and 2', respectively. (b) (c) Function decomposition tree of the cooling system. Although the devices are conceptually different, they can be equated with the way in the function decomposition tree. (b) is the function-first perspective and (c) is the way-first perspective.
}
\label{fig:firstFunctionDecompotion}

\end{figure}

In the conventional function decomposition trees, the decomposition way and the specialization way are combined to form ways for function decomposition, where functions are written as oval functional nodes, the ways are indicated with squares, and the relationship between the functions and ways are represented with connecting lines, as depicted in Figure~\ref{fig:firstFunctionDecompotion} (a). In addition, the execution (achievement) of child functions obtained from a decomposition proceeds from left to right in the function decomposition tree.

The difference between the two decomposition methods explained in Table~\ref{tag:method} are further explained: the decomposition method \#1 exhibits a corresponding implementation, whereas the \#2 does not. In the decomposition tree, the \#1 decomposes the parent function into multiple functional nodes, whereas \#2 replaces the parent function with a single specialized functional node based on the is-a relation. Therefore, the two methods are significantly different from each other, as shown in Figure~\ref{fig:firstFunctionDecompotion} (a).

The resulting diagram is a hierarchical pyramidal decomposition tree structure that decomposed the parent function into multiple subfunctions. This decomposition expression is known as a function decomposition tree.

The parent function can be decomposed by multiple ways. A function decomposition tree comprising the decomposition of the parent function by multiple ways is called a \textbf{general function decomposition tree}, where each decomposition corresponds to a design-alternative representing how to achieve the parent function. In this study, the general decomposition tree has been referred to as the function decomposition tree, as the decomposition of functions by multiple ways is common. 

Function decomposition trees of the device presented in Figure~\ref{fig:heatExchanger} are depicted in Figure~\ref{fig:firstFunctionDecompotion} (b) or (c), which use the methods explained in Table~\ref{tag:method} \#1 to achieve "cool a warm substance" by two different implementation ways: the contact heat exchanger uses the heat conduction way as its implementation way, and the thermal radiator uses the thermal radiation way as its implementation way. To illustrate the expression of function decomposition in line with Figure~\ref{fig:firstFunctionDecompotion} (b). The heat conduction way is achieved by "Make the warm substance existent", "Make the cold substance existent", "Contact these substances", and "Transfer heat". Similarly, the thermal radiation way is achieved by "Make a warm substance existent", "Expose the surface", and "Radiate heat", as shown in  Figure~\ref{fig:firstFunctionDecompotion} (b).

\subsection{Function and action decomposition\label{sec:funcaction}}
In actions, an \textbf{action decomposition tree} \cite{REFCHARM,REFLifeStyle1,REFLifeStyle2,REFLifestyle3} were developed because a virtual device acting on object attributes can be considered to obtain the desired results. The functions and actions can be connected without any gaps because both are expressed in the decomposition trees. Henceforth, we will refer to both decomposition trees as function decomposition trees without distinguishing between them.

Function and action decomposition trees significantly expand the scope of application of the function decomposition. For instance, computer programs are methods for problem solving. They can be considered as artefact devices to imitate the functions of artefacts with equations initiated from principles, or action procedures such as solving equations as directed by humans. Either way, the computer programs can be systematically expressed with a function and action decomposition tree that renders the objective structure explicit.

\subsection{Perspectives and issues in conventional function decomposition\label{sec:convfuncdecomp}}
The perspectives and problems are explained with the conventional function decomposition prior to proposing the new method.

\subsubsection{Two perspectives of conventional function decomposition\label{sec:twoperspective}}

In the conventional function decomposition tree, the schemes of function decomposition can be divided into two major perspectives, which will be explained as follows using Figure~\ref{fig:firstFunctionDecompotion} (b) and (c) as an example.
Figure~\ref{fig:firstFunctionDecompotion} (b) initially places the way nodes and places the functional nodes underneath the way nodes from left to right in the processing sequence. This is referred as the \textbf{way-first perspective}.
Figure~\ref{fig:firstFunctionDecompotion} (c) initially arranges the functional nodes from left to right according to the processing sequence shown in Figure~\ref{fig:heatExchanger}. This is referred as the \textbf{function-first perspective}. (Thereafter the figure arranges the nodes of possible ways.)
There are multiple ways of representing the function decomposition tree, even for schemes as simple as that presented in Figure~\ref{fig:heatExchanger}. In the traditional approach, the skilled personnel take advantage of this representational flexibility and select a systematic representation from those shown in Figure~\ref{fig:firstFunctionDecompotion} for easy comprehension.

\subsubsection{Problems with conventional function decomposition trees\label{sec:problemConvfuncDeomp}}

Though the usefulness of the function decomposition tree has been extensively evaluated through real-world applications\cite{REFCHARM,REFFBRL,REFFunctionalConceptOntology}, describing a proper decomposition tree is challenging for domain experts. The present authors believe that the conventional function decomposition trees have the following drawbacks. 

a. Unfamiliarity with function decomposition trees
In general, researchers aren't accustomed in creating a purpose-based decomposition algorithm. Additionally, novices are often unsure of wording the sentence describing the functions.

b. Too much emphasis on the action of the device
Because the function decomposition is white-boxing, the emphasis is placed on the action of the device though of the way node. Thus, there is no direct relationship between the object-attribute flowchart and the function decomposition trees, and there can be an overlap between the representations of the higher-level functions and the lower-level functions.

c. Processing sequence
The functional nodes that have no processing sequence among them cannot be described. Although there is always a processing sequence from left to right for functional nodes in the function decomposition trees, there is often no processing sequence in decomposing functions in case of multiple inputs to the device, as depicted in Figure~\ref{fig:heatExchanger} (b) and (c). Thus, as shown in Figure~\ref{fig:firstFunctionDecompotion} (b) and (c), the creator of the decomposition tree must "generate" the processing sequence, where the warm material is introduced prior to the cold material.
These issues significantly challenge novices to generate function decomposition trees. However, the problems caused by the proficiency level can be solved with the automatic conversion of the object-attribute flowcharts to the function decomposition trees. 

\subsection{The verb "obtain"\label{sec:verb}}
As functions are expressed using verbs, there must be a common understanding of the meaning of the verbs; otherwise, communication will not be possible. A systematic study of vocabulary was conducted in previous research for functional expression (function vocabulary) in \cite{REFFrameworkFunctionKnowledge}. This section describes (defines) the verb "obtain" and the related verbs most often used to describe the attributes of objects, which are the outputs of devices (ways). This allows a clear distinction between modifications in the attributes of the objects and that in the objects itself.
"Obtain" is used to mean "to make an object (attribute) satisfy a requirement that was not at hand before". For instance, the function of Figure~\ref{fig:heatExchanger} (c) "cool (a warm substance)" could be rendered as "obtain a cold substance (from a warm substance)". As shown in Table~\ref{tag:obtain}, the function "obtain" can be defined recursively using the functions "select," "construct," and "generate."

\begin{table}
\caption{
 "Obtain" is defined cyclically from "Select", "Construct", and "Generate."
}
\hrulefill \\
\begin{itemize}

\item \textit{Obtain}

\begin{itemize}
 \setlength{\parskip}{-1cm} 
  \setlength{\itemsep}{0mm} 
\item \textit{Select}
  \begin{itemize}

	\item No evaluation way
        \begin{enumerate}
	  \item  Randomly select one candidate
        \end{enumerate}
	\item Evaluation way
        \begin{enumerate}
	\item Select or construct evaluation criteria $<$recursion$>$ and evaluate (corresponding to generate evaluation values). \\
	\item  Select one with reference to the evaluation value.\\
        \end{enumerate}
  \end{itemize}

\item \textit{Construct} (parts = operable objects)\\
   \begin{itemize}
   \item Creation way
        \begin{enumerate}
	\item  Obtain parts $<$recursion$>$\\
	\item  "Deform" or "disassemble" the part\\
	\item  Combine parts (select only one)\\
        \end{enumerate}
   \end{itemize}
\item \textit{Generate}
  \begin{itemize}
  \item Generation way
     \begin{enumerate}
     \item 	Obtain the output of the constructed parts\\
     \end{enumerate}
  \end{itemize}

\end{itemize}
\end{itemize}

\hrulefill
\label{tag:obtain}

\end{table}

Subsequently, the parts are defined herein. For artefacts, the parts refer to gears, screws, and pipes that pass through a medium. For algorithms, parts refer to function subroutines as parts to be combined. 

Both "select" and "construct" are used to "obtain" a new part from an existing part. "Select" means picking up an object from the candidates at random or according to the selection criteria that could be selected from the candidates finding a recursion. "Construct" refers to obtaining parts (operable objects) by transforming or disassembling them, and subsequently transforming, decomposing, and combining them to obtain a new part for the purpose. For instance, several sub-parts can be combined into a complete part with "construct".

Let us illustrate the word "generate" with another example. Optimization of the efficiency of GaN LEDs is done using machine learning \cite{REFSimulationbasedMachineLearning}. "Generate" means to essentially create something out of nothing in one view and is used to bring forth a new object of different types from the input. The meaning of different types from the input is explained below in terms of the artefact function. LED lights are artefacts and devices that convert electrical energy into visible light. (Although a portion of the electrical energy is converted into heat as well, heat generation was not considered as a function as it was not the purpose of the device herein.) The input object of this device is electrical energy, and the output object is visible light. In the perspective of existence or non-existence in the category of visible light, something is created out of nothing through the device, and this type of transformation is described as "generate (light)".

Next, we will provide an example of a regression model.\cite{REFPatternRecognition} The linear model with an intercept is written as $f(x)=\sum_i c_i x_i+c_0$ and is "constructed" because they are made of already existing parts, $x_i$ and $c_i$, which are explanatory variables and scalars. In contrast, the values of coefficients $c_i$ are "generated" as a result of training based on training data. It is because the values of coefficients in the trained model have different types from the training data and do not exist before training. Note that the selection of "construct" or "create" depends on the context of describing algorithms. For example, LASSO, which employs the L1 penalty term, was newly discovered in 1986,\cite{REFLinearInversion} i.e. was "created", but it was "constructed" in the sense that the power of the penalty term in the optimization function was changed from the Ridge regression, which uses the L2 penalty term in 1977.\cite{REFRidgeRegression}

Lastly, note that "construct a part and generating an output from the part" may be collectively described as "obtain", expecting that there will be no misunderstanding as the decomposition tree would become redundant. The above minimal definitions will allow us to express "obtain" and the changes in attribute without any misunderstanding.

\section{New function decomposition method\label{sec:newfuncdeomp}}
\subsection{Decomposition methods with new types of functional nodes\label{sec:newtype}}

In this section, a new function decomposition way is proposed. Revisit the correspondence between the object/attribute flowchart and the function decomposition tree. Although the changed attributes constitute the purpose (output) of the object-attribute flowchart, the function generally represents the changes in attributes as well as the action that causes the change. For instance, the objective (object-attribute) in Figure~\ref{fig:heatExchanger} (c) is the cold substance1, whereas the main purpose in Figure~\ref{fig:firstFunctionDecompotion} (b) is to "cool (warm) substance". The "cool" action makes the warm state cold. This would be the main cause of difficulty and confusion faced by beginners during the creation of decomposition trees. The current study solves this problem by introducing a new perspective.

\begin{figure}
\includegraphics[width=8cm]{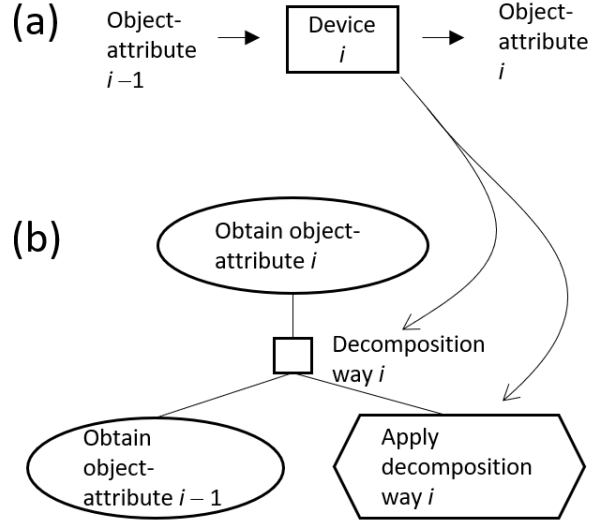}

\caption{  (a) Object-attribute flowchart consisting of object-attribute i?1, device i, and the object-attribute i. (b) Causality-first function decomposition corresponding to (a) is shown using attribute-functional nodes and way-application functional nodes.  The devices in the object-attribute flowchart correspond to the decomposition way in the function decomposition.}
\label{fig:causalityfirst}
\end{figure}

\subsubsection{Two Types of Functional Nodes\label{sec:twotype}}

In contrast to the single type of functional node present in the conventional function decomposition trees, the current study introduces two types of functional nodes that can be distinguished based on their roles in the object-attribute flowchart. As shown in Figure~\ref{fig:causalityfirst}, the ellipsoidal \textbf{attribute-functional node} indicates the change in the object-attribute, and the hexagonal \textbf{way-application functional node} indicates the effect of the device on the object-attribute. The attribute-functional node of the function decomposition corresponds to the object-attribute of the device, which was converted from the object-attribute to "obtain the object-attribute". On the contrary, the way-application functional node corresponds to the action of the device and was converted using the decomposition way to "apply the decomposition way" (the device and way are detailed in Section~\ref{sec:funtiondepintro}).

\subsubsection{Decomposition into Sub-function Sequences in Decomposition Way\label{sec:DecompositionDecomposition}}

The object-attribute flowchart in Figure~\ref{fig:causalityfirst} (a) was converted to that in Figure~\ref{fig:causalityfirst} (b) with the following steps. The object-attribute $i$ was converted to the attribute-functional node "obtain object-attribute $i$", where $i$ indicates the $i$-th term. The device $i$ was converted to the decomposition way $i$, and the object-attribute functional node "obtain object-attribute $i$" was achieved by the decomposition way $i$. The decomposition way $i$ was achieved by the attribute-functional node "obtain the object-attribute $i - 1$" and the way-application functional node "apply the decomposition way $i$", which corresponds to the action of device $i$ and was repeated with respect to $i$.

As represented in Figure~\ref{fig:causalityfirst} (b), "obtain" and "apply" are the most common forms of verbs used in this study because an appropriate automatic conversion in natural language is almost impossible for authors. However, the text can be modified to make the concrete descriptions more natural.

\begin{figure}
\includegraphics[width=12cm]{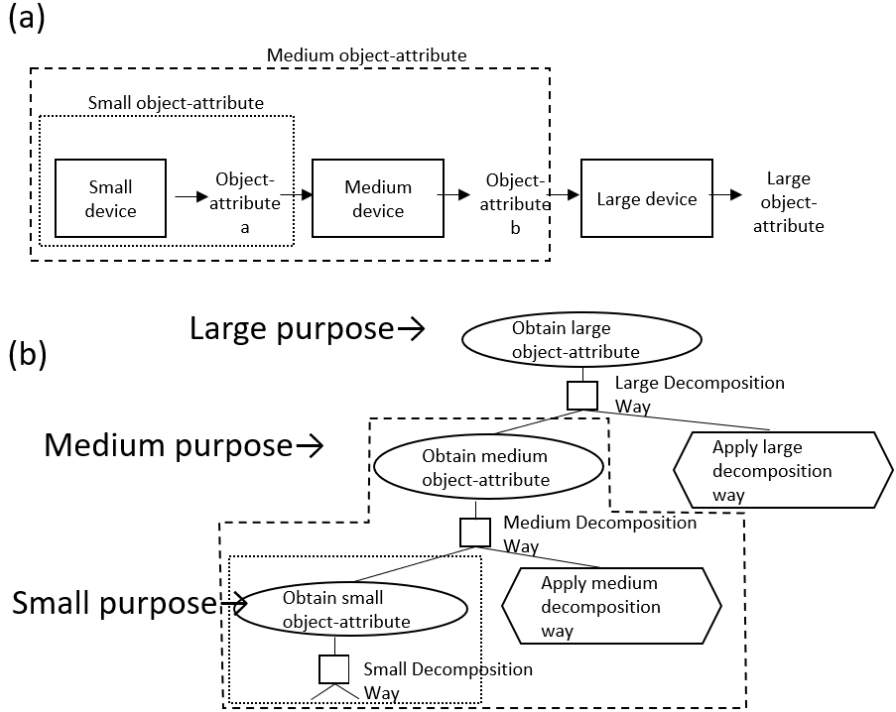}

\caption{
 The object-attribute flowchart showing (a) small device to large object-attributes and (b) its causality-first function decomposition.
}
\label{fig:conversion}
\end{figure}

How can we observe this function decomposition translating for the sequence of object-attributes and devices shown in Figure~\ref{fig:conversion} (a)? This function decomposition can be regarded as forming a nested structure with object-attributes, where devices and ways were categorized as large, medium, and small. In particular, the medium object-attribute depicted in Figure~\ref{fig:conversion} (a) was converted into the large object-attribute using the large device, corresponding to the large decomposition way. The medium object-attribute exhibited an internal structure, whereas the small object-attribute exhibited a structure transformed into the object-attribute b in the medium device, corresponding to the medium decomposition way. This can be considered as a perspective that clearly depicts the present attribute (denoted as large object-attribute) as a result of attribute changes involving the nested past, i.e., the result of a causal chain. Thus, this new proposed method was termed as causality-first perspective of function decomposition (or \textbf{causality-first function decomposition}). Additionally, this sequence of function decomposition can be stated as \textbf{function decomposition chain}.

This decomposition scheme constructs a network of chains rather than a decomposition tree. Thus, we will call it a causality-first function decomposition network (for simplicity, a \textbf{function decomposition network}).

\subsubsection{Specialization Way Decomposition\label{sec:specializationway}}

The conversion corresponding to the specialization way decomposition (\#2 in Table~\ref{tag:method}) was performed according to the conventional function decomposition trees. The functional nodes used in this decomposition were only attribute-functional nodes. In addition, there are no way-application functional nodes in this conversion (the same as Figure~\ref{fig:firstFunctionDecompotion} (a)), because there is no corresponding device for the specialization way.

In conventional function decomposition trees, the distinction between the decomposition way and the specialization way is ambiguous. Because those who write function decomposition trees also understand ontology, these distinctions have been made implicitly, but they should be explained explicitly, especially for novices.

\subsection{Features of function decomposition tree in causality-first perspective\label{sec:featuresofcausality}}
\subsubsection{Role Assignments for Functional Nodes\label{sec:RoleAssignments}}

Although the conventional function decomposition is always aware of the conversion of the input to the output object-attributes for the device, the conversion of function decomposition trees decomposed in terms of paraphrased functions. On the contrary, the expression of the device action in the causality-first perspective is represented by the role of the way-application functional node, and the attribute-functional nodes describe only the object-attributes achieved by the ways. 

\subsubsection{White-boxing of Decomposition Way\label{sec:Whiteboxing}}
\begin{figure}
\includegraphics[width=12cm]{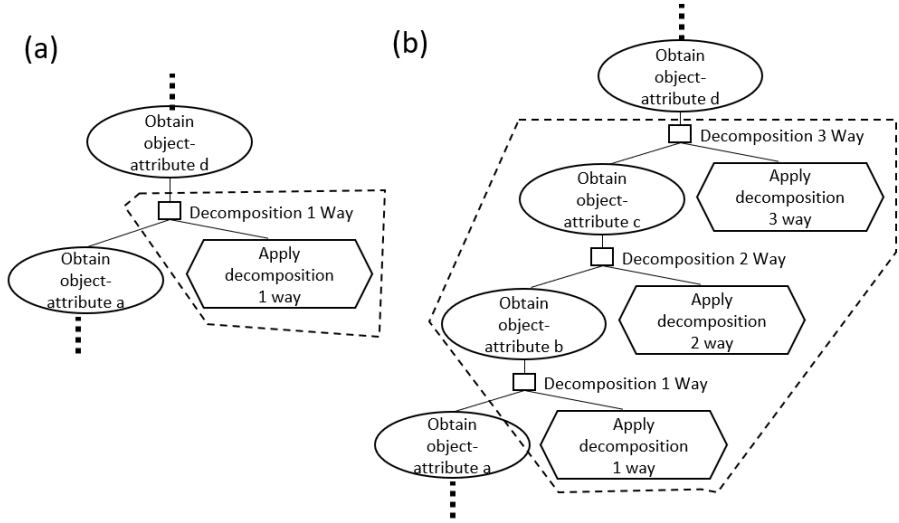}
\caption{
 White-boxing of decomposition way 1 in (a) was conducted with the insertion of the function decomposition chain in (b).
}
\label{fig:whiteboxing}
\end{figure}

The way-application functional nodes are not functionally decomposed in the causality-first perspective. Instead, white-boxing of the decomposition way was achieved in the causality-first function decomposition by inserting function decomposition chains in the longitudinal (causal) direction, as depicted in Figure~\ref{fig:whiteboxing}.

\subsubsection{Network Connection of Function decomposition Chains\label{sec:NetworkConnection}}
\begin{figure}
\includegraphics[width=12cm]{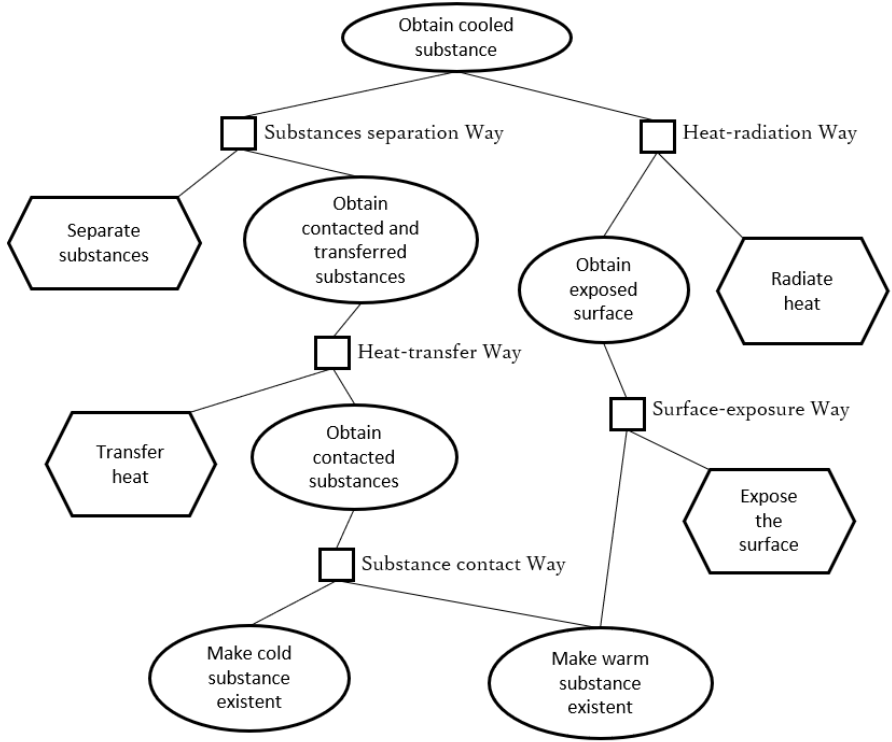}
\caption{
 Function decomposition network with the causality-first perspective corresponding to Figure~\ref{fig:firstFunctionDecompotion} (b) and (d). 
}
\label{fig:network}
\end{figure}

The object-attribute flowcharts in Figure~\ref{fig:firstFunctionDecompotion} (b) and (d) can be converted according to the causality-first conversion rules in Figure~\ref{fig:causalityfirst}. An example of the transformation of Figure~\ref{fig:heatExchanger} (b) (d) into a function decomposition network is shown in Figure~\ref{fig:network}.

The relationship between the function decomposition chain and the function decomposition network is equivalent to the relationship between the function decomposition tree and the general function decomposition tree.\cite{REFOntologicalModel} In addition, the same multiple ways (referring to both decomposition and specialization ways) were represented as design alternatives along with a relationship that depicts a function achieved by satisfying at least one way.

\subsubsection{Uniqueness of Function Decomposition Network\label{sec:UniquenessFunction}}

The object-attribute function flowchart was uniquely converted to the function decomposition network. Therefore, how to connect function and way nodes are also determined uniquely. Existing decomposition network is preserved when new decomposition chains is added. In other words, new decomposition chains can be added at any position. This is exemplified in the Electronegativity portion of Section~\ref{sec:Superposingothergoals}.

\subsubsection{Description of Independently Achievable Procedures\label{sec:DescriptionIndependently}}

The processing sequence only executes the way-application functional node after all the attribute-functional nodes have been achieved. In this context, the achievement of all attribute-functional nodes serves as a trigger for the execution (\textbf{execution trigger}) of the way-application functional nodes. Thus, the rules of processing sequence in the function decomposition tree (left to right) were able to be abolished in the function decomposition network, and the only sequence of processing immediately below the way $i$ was to perform the way-application functional node $i$ at the end.

\subsection{Computer-aided generation of causality-first function decomposition networks\label{sec:Computeraided}}

With the explanations given so far, the conversion rules from the object-attribute flowchart to the function decomposition network should be obvious without the further explanation. This sub-section introduces a software to automatically generate function decomposition networks.

As beginners face difficulty in directly performing function decomposition, the sequential algorithms are first prepared with Cytoscape GUI-based network visualization software.\cite{REFCytoscape} Then, it is converted into a function decomposition network using the Python computer program we developed.\cite{REFCytoscapeHelper} The output network is in the Graphviz or Cytoscape CX format and the latter can be edited with Cytoscape GUI.

\begin{figure}
\includegraphics[width=12cm]{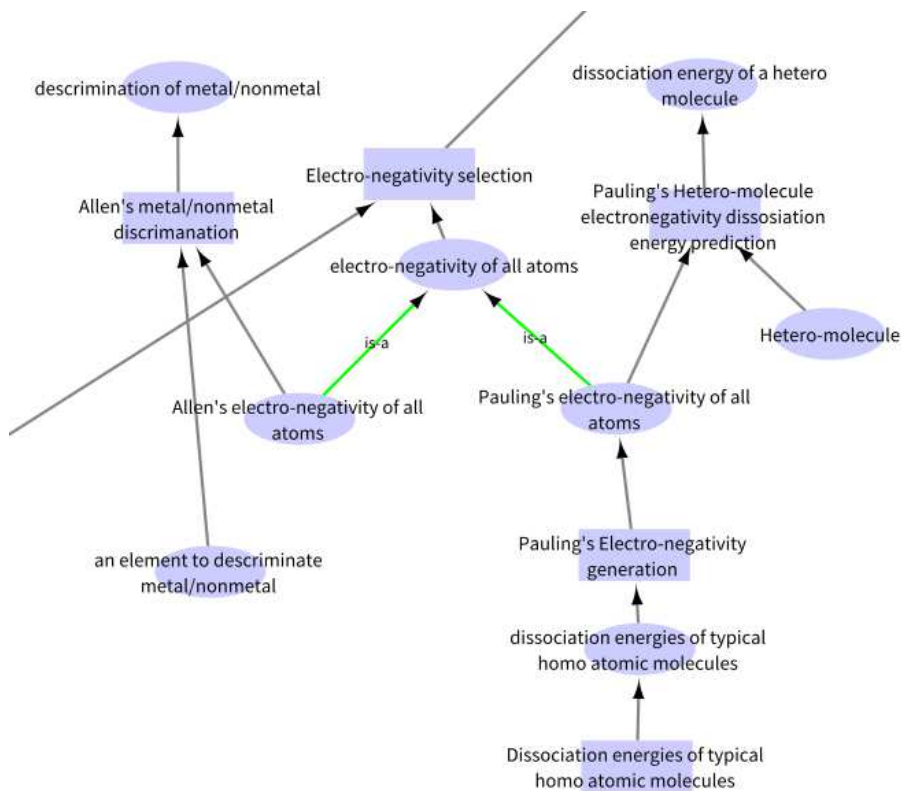}

\caption{
The device-based workflow for the electronegativity generation portion was cut from the overall workflow. Lines that are present in the original diagram but not required are not shown.
}
\label{fig:electronegativity}
\end{figure}

We explain how to describe an object-attribute flowchart based on the sample/descriptorTarget in the repository.\cite{REFCytoscapeHelper} Figure~\ref{fig:electronegativity} shows a device-based workflow associated with Electronegativity. The object-attribute function nodes are shown as ovals, the device names are represented as squares, and their input-output relationships are represented by arrows. In addition, the "is-a" label is added to the arrows to express the is-a relationship between the attribute function nodes at the same time. The former corresponds to Table~\ref{tag:method}, \#1 and the latter to Table~\ref{tag:method}, \#2.

After saving in a CX JSON file in Cytoscape, the command,
\begin{lstlisting}[basicstyle=\ttfamily\footnotesize, frame=single][htbp]
$ python cx_network_main.py --graphviz_FDN file1 file2 ...
\end{lstlisting}
is used to create the function decomposition network using graphviz \cite{REFGraphvizDynagraph}, where file1, file2,... are the Cytoscape CX JSON format files. The command, 
\begin{lstlisting}[basicstyle=\ttfamily\footnotesize, frame=single][htbp]
$ python cx_network_main.py --cx_FDN file1 file2 ...,
\end{lstlisting}
creates the function decomposition network in Cytoscape CX format. It is noted that arrows are added to correspond to the workflow order.

The workflow in sample/descriptorTarget can be automatically converted to function decomposition network in Graphviz format, and the resulting network is shown in Figure~\ref{fig:heatExchanger} of \cite{REFSMinZenodo}. The function decomposition network corresponding to Figure~\ref{fig:electronegativity} is shown in Figure~\ref{fig:firstFunctionDecompotion} of \cite{REFSMinZenodo}.

For the problem that the expression of the function is unfamiliar to inexperienced users in 2.4.2a, a minimum expression can be obtained by designating the attribute-functional node as "obtain the object-attribute" and that for the way-application node as "apply the way". The overlap of representation problem doesn't occur as a result of automatic transformation when there is no overlap among the object-attribute names and among the way names in the object-attribute flowchart. Thus, novices can make function decomposition networks.

\section{Application 1: systematic explanation of materials informatics\label{sec:Applicationsystematicexplanation}}

In this section, we analyze functions where descriptors are generated from a crystal and are related to crystal target variables, as an example of materials informatics in the function decomposition network. Descriptors and their relationship to the target variables are explained in conventional text (, or sequential algorithm) in Section~\ref{sec:sequentialandpurposedecomp} of \cite{REFSMinZenodo}. The decomposition and specialization ways are indicated by open squares and grey-filled squares, respectively. In addition, certain nodes were omitted and rearranged for appropriate viewing. Function decomposition network with the causality-first perspective is provided in \cite{REFSMinZenodo}.

\begin{figure}
\includegraphics[width=12cm]{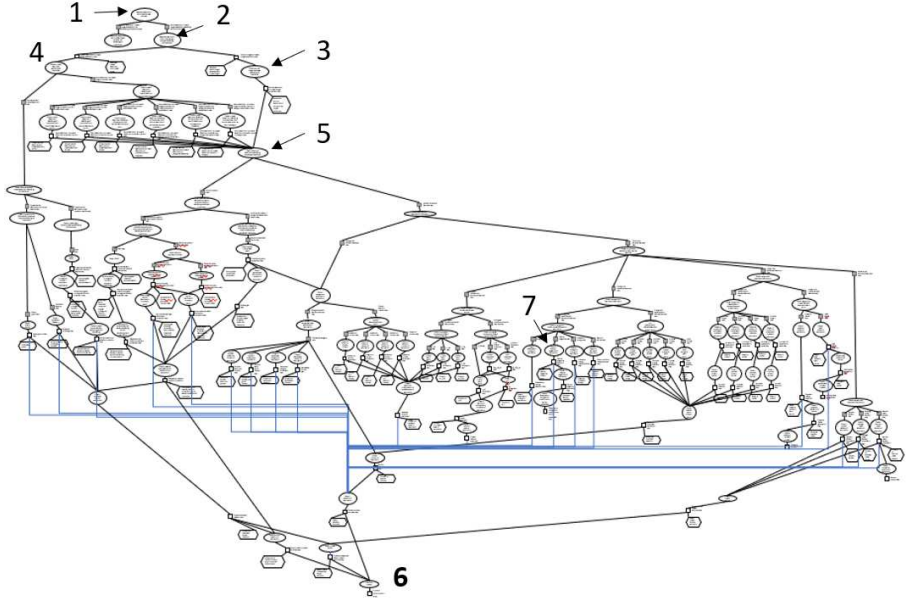}

\caption{
 The Overall function decomposition network achieving the top goal: "Associate descriptors with crystal objective variables" (1 in the figure) from the bottom node "obtain a crystal" (6). The numbers in the figure indicate the same position in subsequent figures.
}
\label{fig:overallCrystal}
\end{figure}

\subsection{Overall feature\label{sec:Overallfeature}}

The topmost purpose was set as "Associate descriptors with crystal objective variables" (1) in Figure~\ref{fig:overallCrystal}. Ways between the topmost purpose and the bottom-line purpose "get a crystal" (6) were white boxed systematically. The figure was first written with an object-attribute flowchart and the decomposition by specialization way, then automatically converted into function decomposition network. Lastly, each node was modified with an appropriate English expression.

\begin{figure}
\includegraphics[width=15cm]{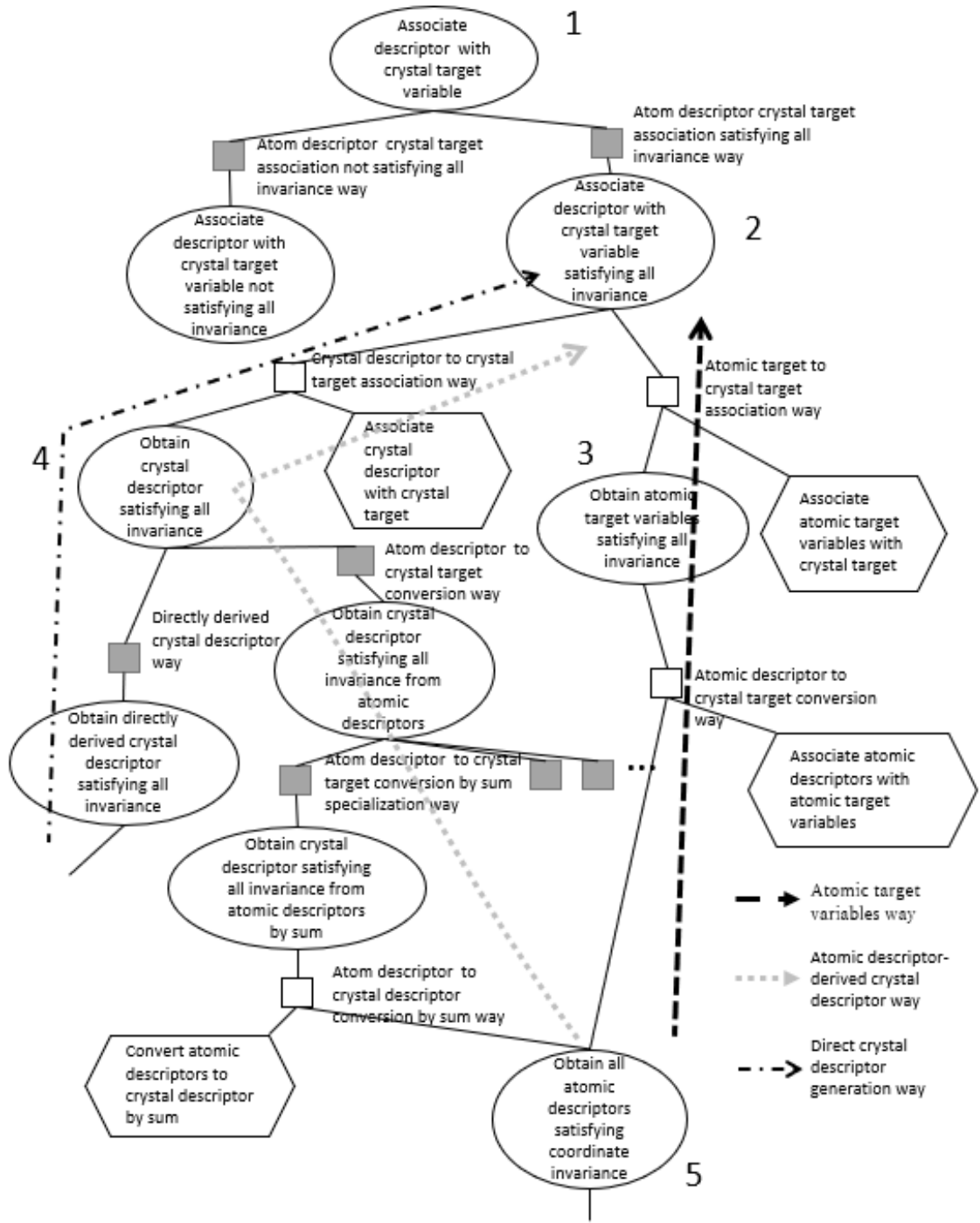}

\caption{
Enlarged view in the part of 1--5 nodes of Figure~\ref{fig:overallCrystal}; certain nodes were omitted and rearranged for appropriate viewing. Arrows show the three ways; the atomic target variable way, atomic-descriptor derived crystal descriptor way, and direct crystal descriptor generation way, which are explained in \cite{REFSMinZenodo}. The numbers in the figure correspond to those in Figure~\ref{fig:overallCrystal}.
}
\label{fig:enlargedoverall}
\end{figure}

Enlarged view in the part of 1 -- 5 nodes of Figure~\ref{fig:overallCrystal} is shown in Figure~\ref{fig:enlargedoverall}; some nodes were omitted and rearranged for appropriate viewing. There are three ways to achieve the major goal, each indicated by an arrow. They are named the atomic target variable way, atomic-descriptor derived crystal descriptor way, and direct crystal descriptor generation way, the number of which correspond to those in Figure~\ref{fig:overallCrystal}. They are also explained as textual descriptions in \cite{REFSMinZenodo}. 

\subsection{Superposing other goals\label{sec:Superposingothergoals}}

In conventional function decomposition trees, it is sometimes necessary to rewrite the function decomposition tree structures when a new goal is added to an existing decomposition tree. This is not only a case of polishing the function decomposition tree, but also a case of correcting the essential problem of fixing an inter-node order that differs if new nodes are added as described in Section~\ref{sec:specializationway} . We show that new goals can be easily added to a function decomposition network.
Electronegativity is a relative measure of the strength with which atoms in a molecule attract electrons and is used as one of the descriptors of atoms. However, it was originally invented as an internal measure to predict the dissociation energy of heteromolecules from homomolecules.\cite{REFGraphvizDynagraph}

\begin{figure}
\includegraphics[width=12cm]{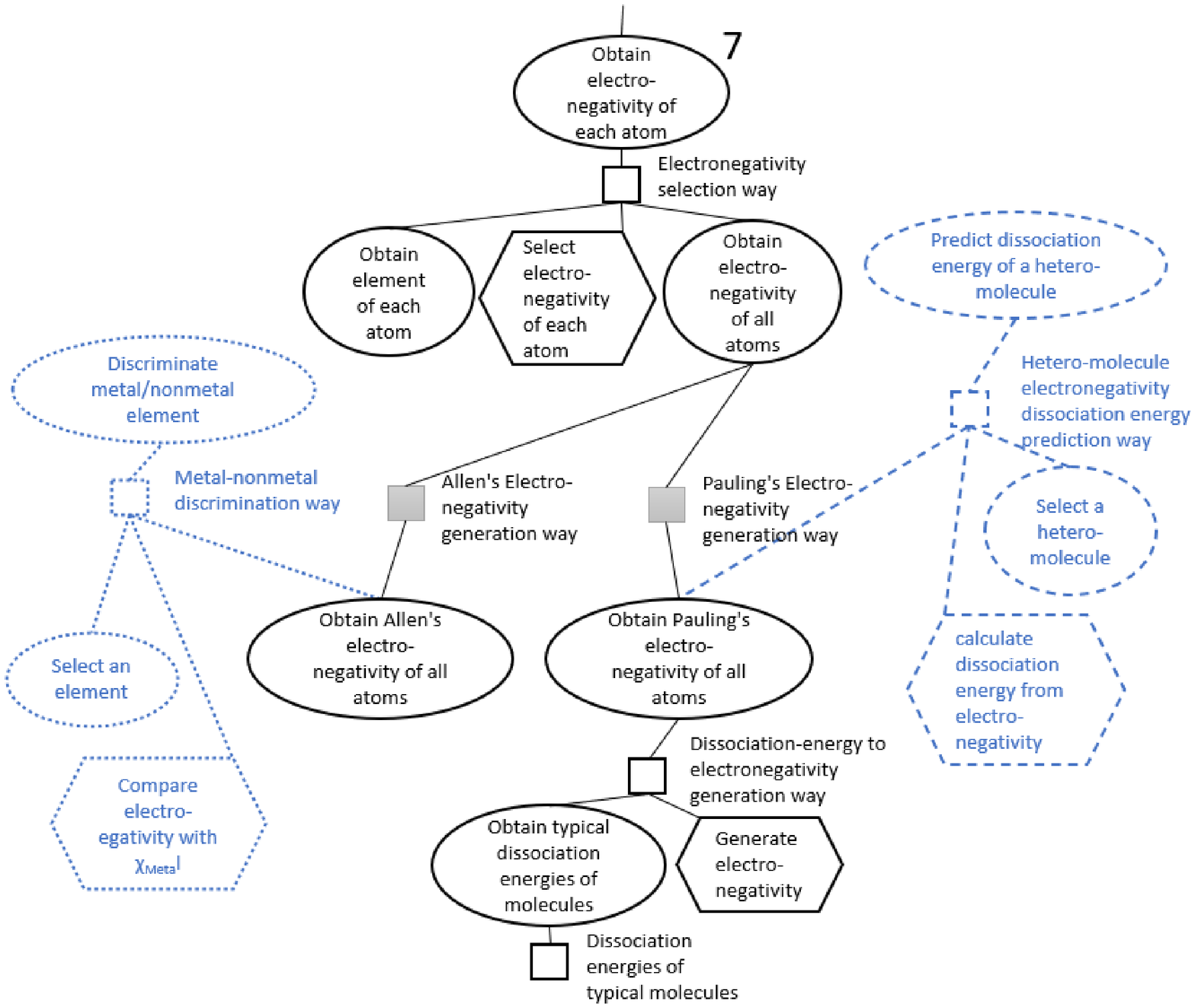}

\caption{
Enlarged view of parts 7 of Figure~\ref{fig:overallCrystal}. The function decomposition chains to obtain the electronegativity are presented. Pauling's original final goal, "Predicting dissociation energies of different molecules" in electronegativity, is superposed by the nodes using dashed shapes and dashed lines as another function decomposition chain. The goal, "Discriminate metal/nonmetal element" is also added as dotted lines and shapes. 
}
\label{fig:enlargedpartSeven}
\end{figure}

Function decomposition networks can superpose these different targets on the existing network. For example, Pauling's original final goal, "Predicting dissociation energies of different molecules" in electronegativity, is superposed by a dashed line in another chain of function decomposition networks in Figure~\ref{fig:enlargedpartSeven}. (Dashed blue lines are used for clarity in the additions.) Furthermore, it is known that Allred's electronegativity \cite{REFElectroNegativityAllred} indicates the nature of metals and non-metals to some extent. Figure~\ref{fig:enlargedpartSeven} also shows his goal, i.e., "Discriminate metal/nonmetal element". It also illustrates the name; the electronegativity is the same, but their generation ways are different. As with the citation network of articles, concepts and metrics that have been identified as useful will be shown to be studied for different goals (functions) and variants (ways) in function decomposition. Note that research on the electronegativity measure started in 1931's, but is still ongoing.\cite{REFElectroNegativityRahm,REFThermoChemical}.

\subsection{Function decomposition chains and design-alternative considerations\label{sec:Functiondecompositionchains}}

\begin{figure}
\includegraphics[width=12cm]{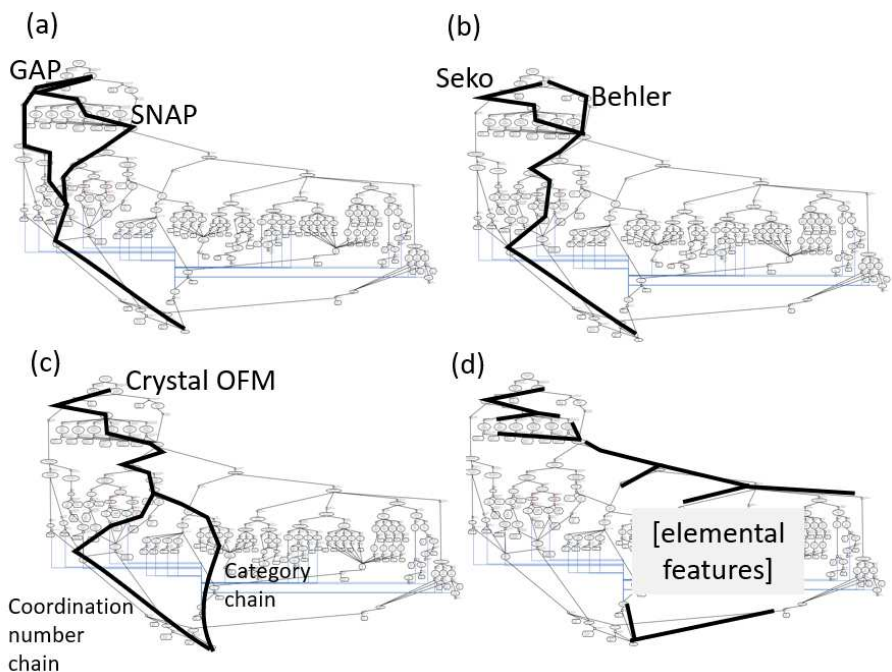}

\caption{
 (a) Function decomposition chains of SOAP-GAP and SNAP. (b) Chains of Behler and Seko et al. (b) The coordinate number and category decomposition chains constituting crystal-OFM. (d) Function decomposition chain using elemental descriptors.
}
\label{fig:SOAPGAP}
\end{figure}

Finally, the methods used in smooth overlap of atomic positions kernel and the Gaussian approximation potential (SOAP-GAP) \cite{REFGaussianApproximationPotential}, spectral neighbor analysis potential  (SNAP) \cite{REFSpectralNeighborAnasysis},  Behler et al \cite{REFBehler}, Seko et al.\cite{REFSeko}, orbital field matrix (OFM) \cite{REFOFM}, and regression with elemental features are visualized in Figure~\ref{fig:SOAPGAP}. The textual details of each will be placed in \cite{REFSMinZenodo}.

In Figure~\ref{fig:SOAPGAP} (a), SOAP-GAP regresses the crystal target variable by directly obtaining the SOAP crystal descriptor from the extended atomic positions, whereas SNAP interprets SOAP as atomic descriptors and regresses the crystal target variable by adding them to the crystal descriptor. Moreover, the chain in (b) starts from the same atomic distribution, but Behler et al. use the symmetry function as an atomic descriptor to associate the atomic and the crystal target variable. Although Seko et al. use the same symmetry function, it combines with linear regression to transform the crystal descriptor by summation and regress the crystal target variable.

In (c), OFM uses generalized coordination numbers calculated from the relative atomic positions along with the categorical features converted from the elements to produce the matrix representation of the central and neighboring atomic environments. In addition, other category variables are shown in the function decomposition network, and the possibility of using other category variables can also be considered.

In (d), the generation of atomic descriptors were performed with the elemental features that are non-atomic position descriptors. In addition, the atomic descriptors were converted to a crystal descriptor by several ways to transform crystal descriptors from atomic descriptors with atom-reordering invariance. The decomposition by specialization way of elemental features was performed using the type of variables and dimension of the features. The purposes of the elemental features may be able be explained more clearly by increasing the specialization decomposition.

It is difficult for a non-specialist to understand the whole theories of a given field. On the other hand, the differences between the methods in Figure~\ref{fig:SOAPGAP} can be hoped to reduce efforts required by, for example, informatics researchers to understand the challenges/problems of materials science. This would be one of the major advantages of the function decomposition networks.

Finally, we provide additional insights from these analyses using function decomposition networks. The methods of SOAP-GAP, SNAP, and Seko et al. selected regression models depending how the crystal descriptor is associated with the crystal target variable. Behler et al.'s method is different from the others because they defined the crystal target variable as the sum of the atomic target variable. Note that only the summation way facilitates the conversion to atomic density by considering the total energy as a function of atomic density, as used by SOAP-GAP, SNAP, Seko et al., and OFMs. However, there exists the other ways between the function attribute nodes, "obtain crystal descriptor satisfying all invariance" and "obtain all atomic descriptor satisfying coordinate invariance" in Figure~\ref{fig:enlargedoverall}. The other conversion ways such as standard deviation may contribute to the improvement of the regression performance.

\section{Application 2: check for consistency of subroutines in computer program libraries\label{Applicationcheckconsistency}}

In recent years, computer library distribution through public repositories has become popular in materials informatics. For the library to be widely used, the library documentation needs to be maintained, but before that, creating a library that integrates miscellaneous subroutines developed by multiple authors or by a single author over a long period of time requires a unified naming scheme and class/subclass structure. For this purpose, it is first necessary to grasp the overall picture of the functions and the names of the subroutines. (To avoid confusion with function, which is used as function decomposition, a subroutine is used for a computer subroutine.)

\begin{figure}
\includegraphics[width=10cm]{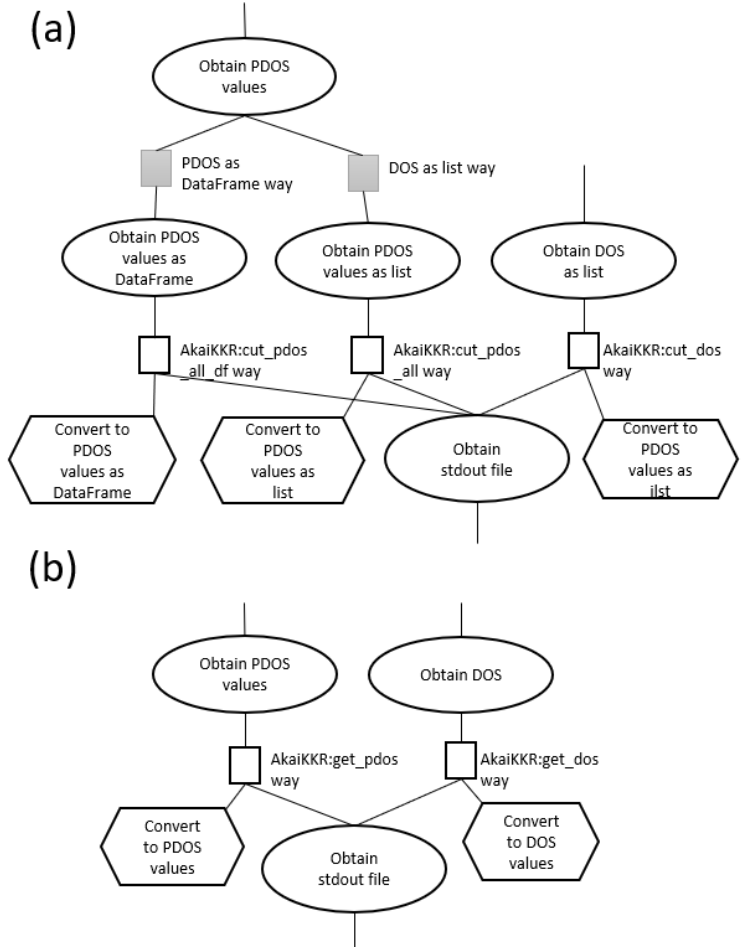}

\caption{
 (a) Initial function decomposition network, (b) modified function decomposition network; "Class name:member function name" is shown in a way node.
}
\label{fig:subroutineInitial}
\end{figure}

Figure~\ref{fig:subroutineInitial} shows a part of the function decomposition network of PyAkaiKKR \cite{REFSearchByAkaiKKR} used in \cite{REFFukushima}. Their functions are to get density of states (DOS) or partial density of states (PDOS) from the stdout file output of the AkaiKKR package\cite{REFKKR}. Figure~\ref{fig:subroutineInitial} (a) shows a network when the miscellaneous subroutines already existed are simply put together. "Class name:member function name" is shown in a way node. It had two problems. One is that many other subroutines have subroutines beginning with "get", which are not shown here, but these have subroutine names beginning with "cut." Second, DOS outputs in list, while PDOS outputs in list and DataFrame\cite{REFPandas} and the subroutines that executes them were created separately. The latter is represented by two decompositions by specialization way in Figure~\ref{fig:subroutineInitial}. They should start with "get" and the output format should be selectable as a subroutine option. The results of these modifications are shown in Figure~\ref{fig:subroutineInitial} (b). The output type is now handled as option in these subroutines. 

\begin{figure}
\includegraphics[width=15cm]{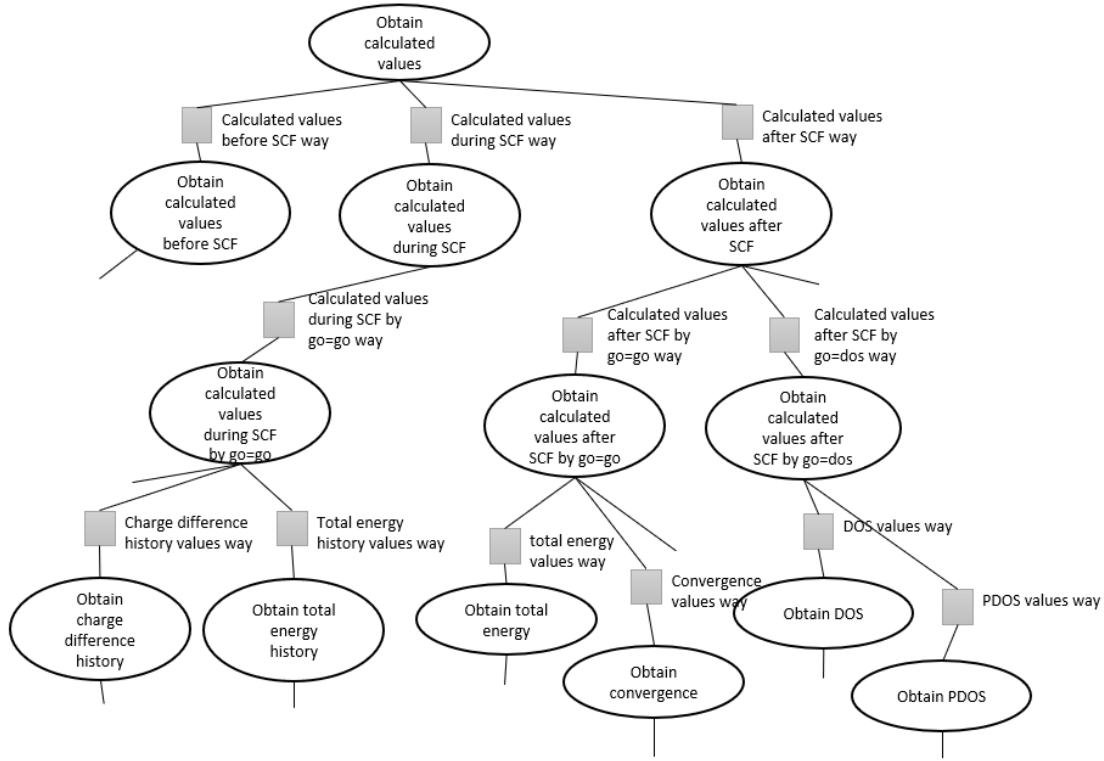}

\caption{
 The "Obtain calculated value" is set as the upper goal and is connected to "obtain DOS", "obtain PDOS", etc. using the decomposition by specialization way.
}
\label{fig:obtainDOS}
\end{figure}

For novices who are not familiar with density functional theory, it is helpful to have it stated at the same time which keywords for which runs yield which physical quantity. This paragraph further categorizes and organizes subroutine to facilitate user understanding using decomposition by specialization way. The higher-level goal of "obtain DOS" in Figure~\ref{fig:subroutineInitial} would be "obtain calculated values after SCF", where SCF is an abbreviation of satisfying the condition of self-consistent field which must be satisfied in electronic state calculations. In AkaiKKR, the calculation is performed with the keyword go="go" to obtain SCF. For example, to calculate DOS, go="dos". There are other physical quantities that can be calculated after SCF, so it would be more comprehensive and user-friendly function decomposition if they are described at the same time. The results of organizing these are shown in Figure~\ref{fig:obtainDOS}. It is possible to find out which value of go will achieve which higher-level goal.

The advantages of organizing the subroutines by function decomposition are that the authors can easily grasp the contents of the libraries, and this reduces hesitation in making changes for unification. By performing decomposition by specialization way at the same time, the contents that had previously been vaguely understood are now easier for the authors to sort out and organize. While other methods can be used to unify name rules, the advantage of function decomposition networks is that special decompositions can be described at the same time.

Similar things are done in nursing guidelines integration occurred in hospital integration \cite{REFCHARM}. This section shows another example of writing a function decomposition network can clarify differences and/or confusion and facilitate the knowledge integration process.

\section{Discussion on function decomposition network\label{sec:Discussionfunctiondecomposition}}
\subsection{Function decomposition with two functional nodes}

The causality-first perspective described states that the action of the device is represented by the role of the way-application functional node, and the attribute functional node considers only the object-attributes achieved using the way, as shown in in Section~\ref{sec:RoleAssignments}. The left-to-right execution order is eliminated, and the achievement of attribute function nodes trigger execution the way-application function node as described in Section~\ref{sec:DescriptionIndependently}. The lack of any irregularities in the function decomposition networks presented in Section~\ref{sec:Applicationsystematicexplanation} and Section~\ref{sec:Applicationsystematicexplanation} confirmed that the proposed methods based on the causality-first perspective performed appropriately. 

\subsection{Developing verb vocabulary}

There must be a common understanding of the meaning of the verbs as explained in Section~\ref{sec:verb}. The minimum verbs needed for description are called the basic verb vocabulary. We categorized "obtain" and explained "select", "construct", and "generate"; along with "apply", these are the most basic vocabulary in materials informatics. Furthermore, in data science, "regress," "classify," and "associate," would be its candidates among other verbs. Since duplicate verb expressions lead to the same function decomposition being performed with several different expressions, the definition of the basic verb vocabulary in materials informatics including examples needs to be established as in the research on lifestyle.\cite{REFLifestyle3} It is also necessary if the algorithms are grouped by computers. 

\subsection{Connection between different function decomposition networks}

Existing decomposition trees may comprise a mixture of function-first and way-first perspectives, and the hierarchical structures may differ between creators though they executed the same white-boxing. Thus, finding similarities/differences among the function decomposition trees is difficult. On the other hand, the function decomposition networks can identify them easily, resulting in simple overall knowledge. This would be more advantageous when using a computer to determine algorithmic differences.

\subsection{Generation of function decomposition network examples}

Although automatic generation of the function decomposition network from the object-attribute flowchart is now possible, the process of manually modifying the verb vocabulary and expressions remains a two-fold effort. However, a direct construction of the decomposition network will reduce the creator's required time and effort by half.

In this study, we constructed our own function decomposition network from an object-attribute flowchart with the is-a specialization description. Thus, examples of one's own domain can be constructed by oneself. Once one becomes familiar with the function decomposition network through the generated examples, the function decomposition network can be directly generated instead of using the object-attribute flowchart, thus reducing the amount of work involved in its generation.

Moreover, the modification or addition of parts to the function decomposition network is relatively easy even if its complete generation from scratch appears difficult. In addition, the use of a personalized function decomposition network makes it easier to add and expand the specific function decomposition network. Thus, a conversion tool would be useful in this aspect.

\subsection{Decomposition by specialization way}

\subsubsection{Merit}
Figure~\ref{fig:enlargedoverall} described that "Obtain crystal descriptor satisfying all invariance from atomic descriptors" can be realized in various ways. The decomposition network allowed us to simultaneously describe these is-a specializations along the manner of their creation. This is expected to lead to a systematic understanding and the discovery of further different ways.

\subsubsection{Demerit}
In Figure~\ref{fig:overallCrystal}, the specialization way was prioritized after trial and error. However, the specialization way is introduced to organize functions, not to describe devices, and there is a room for human thought and the order of the decomposition way, and the specialization way can be changed. Therefore, the introduction of the specialization way decomposition may make the function decomposition network more difficult to understand.

A more sophisticated solution would include an interactive system that raises questions and selects one of the methods during conflict between the decomposition and specialization ways. The implementation of this resolution will be explored in future research.

\subsection{Execution of function decomposition network to ensure correctness}

The conventional function decomposition trees could only be proved correct by a human being. This is because the correspondence with the workflow was not perfect. On the other hand, the function decomposition network can be executed in principle because it can be transformed from the workflow. Then, execution can guarantee the correctness of the function decomposition. This will be the subject of future research.

\subsection{Improvement of the representation problem of conventional function decomposition trees}

There are two types of functional nodes: attribute-functional nodes and way-application functional nodes, and novices have a vague understanding on these. Thus, the selection of a representation of the same attribute-functional node above and below a way could causes duplicate representation (Section~\ref{sec:problemConvfuncDeomp}.b). No duplication occurs if an application-functional node is placed at the end with (a) attribute-functional node(s) that is(are) the input node(s) (execution trigger) of the way node. Furthermore, if more than one functional node corresponding to multiple devices is to be written under a way node as conventional function decomposition trees, the guideline may be to write an attribute functional node first, followed by way-application functional nodes.

As certain function decomposition tree assets \cite{REFCHARM} already exist, rewriting all of them in a function decomposition network would be require excessive human effort. However, in continuously expanding/correcting fields, frontline staffs who are novices in function decomposition would require to add/modify nodes to their existing function decomposition tree assets even if they are novices in the function decomposition. Difficulties in adding or modifying expressions to conventional decomposition trees can be resolved by the guidelines in the previous paragraph, and the existing function decomposition tree assets can be easily maintained on a continuous basis.

\section{Conclusions}

As interdisciplinary research flourishes, there is a need for systematic and easily understood explanatory methods to facilitate communication about each other's research and its problems. In context, function and action decomposition trees have gained reputation as methods for systematic explanation, but they are unwieldy for novices. Thus, this study describes the current problems of function decomposition trees and proposes a new function decomposition network with the causality-first perspective to solve these problems.

In the proposed function decomposition network, the correspondence with the object-attribute flowchart that tracks object-attribute changes was obtained by separating the conventional functional node into an attribute-functional node for function achievement and a way-application functional node for representing the action of the device. The functional and action decomposition network with the causality-first perspective can be transformed from the object-attribute flowchart and specialization descriptions with the is-a relation; the conversion tool was constructed and published. 

As the function decomposition tree/network requires a common understanding of the verb vocabulary, the most used verb "obtain" were defined along with related verbs. A possible method was obtained by performing function decomposition for a materials informatics problem. The function decomposition was also used to organize computer libraries. Lastly, the limitations and prospects of the current study were discussed.

\section*{Acknowledgements}
This work is supported by the Ministry of Education, Culture, Sports, Science, and Technology of Japan (MEXT) ESICMM Grant Number 12016013, the Program for Promoting Research on the Supercomputer Fugaku (DPMSD), the JST-Mirai Program "Development of Materials Design Workﬂow and Data Library for Materials Foundry," Grant Number JPMJMI18G5, JPMJMI21G2 and JSPS KAKENHI Grants 20K05311, JP19H05815 (Grants-in-Aid for Scientiﬁc Research on Innovative Areas Interface Ionics), 20K05068 and JP21H01375 Japan.

\end{document}